\documentclass[runningheads]{llncs}
\usepackage{graphicx}
\usepackage{epstopdf}
\usepackage{microtype}
\usepackage{subfigure}
\usepackage{booktabs} % for professional tables
\usepackage{empheq}
\usepackage[misc]{ifsym}
\usepackage{amsmath,amsfonts,bm}

\def\eqref#1{equation~\ref{#1}}
\def\1{\bm{1}}

\DeclareMathAlphabet{\mathsfit}{\encodingdefault}{\sfdefault}{m}{sl}
\SetMathAlphabet{\mathsfit}{bold}{\encodingdefault}{\sfdefault}{bx}{n}

\newtheorem{theo}{Theorem}
\newtheorem{proo}{Proof}

\begin{document}
\title{Adversarial Invariant Feature Learning with Accuracy Constraint for Domain Generalization}
\titlerunning{Adversarial Invariant Feature Learning with Accuracy Constraint}
\author{Kei Akuzawa\inst{1}\Letter \and
Yusuke Iwasawa\inst{1} \and
Yutaka Matsuo\inst{1}}
\authorrunning{Akuzawa et al.}
\institute{School of Engineering, The University of Tokyo, 7-3-1 Hongo, Bunkyo-ku, Tokyo 113-8656, Japan. \email{\{akuzawa-kei,iwasawa,matsuo\}@weblab.t.u-tokyo.ac.jp}
}
\maketitle              % typeset the header of the contribution
\begin{abstract}
Learning domain-invariant representation is a dominant approach for domain generalization (DG), where we need to build a classifier that is robust toward domain shifts.
However, previous domain-invariance-based methods overlooked the underlying dependency of classes on domains, which is responsible for the trade-off between classification accuracy and domain invariance.
Because the primary purpose of DG is to classify unseen domains rather than the invariance itself, the improvement of the invariance can negatively affect DG performance under this trade-off.
To overcome the problem, this study first expands the analysis of the trade-off by Xie et. al. \cite{Xie+2017}, and provides the notion of {\em accuracy-constrained domain invariance}, which means the maximum domain invariance within a range that does not interfere with accuracy.
We then propose a novel method {\em adversarial feature learning with accuracy constraint}  (AFLAC), which explicitly leads to that invariance on adversarial training.
Empirical validations show that the performance of AFLAC is superior to that of domain-invariance-based methods on both synthetic and three real-world datasets, supporting the importance of considering the dependency and the efficacy of the proposed method.

\end{abstract}

\section{Introduction} \label{intro}

% ドメイン汎化の重要性，タスクの説明
In supervised learning we typically assume that samples are obtained from the same distribution in training and testing; however, because this assumption does not hold in many practical situations it reduces the classification accuracy for the test data \cite{Torralba+2011}.
This motivates research into domain adaptation (DA) \cite{Ganin+2016} and domain generalization (DG) \cite{Blanchard+2011_DG}.
DA methods operate in the setting where we have access to source and (either labeled or unlabeled) target domain data during training, and run some adaptation step to compensate for the domain shift.
DG addresses the harder setting, where we have labeled data from several source domains and collectively exploit them such that the trained system generalizes to target domain data without requiring any access to them.
Such challenges arise in many applications, e.g., hand-writing recognition (where domain shifts are induced by users, \cite{Shankar+2018}), robust speech recognition (by acoustic conditions, \cite{Sriam+2018}), and wearable sensor data interpretation (by users, \cite{Erfani+2016_sensor}).

\begin{figure}[t]
  \centering
  \vspace{-0.3cm} % 応急処置
  \includegraphics[width=12.5cm]{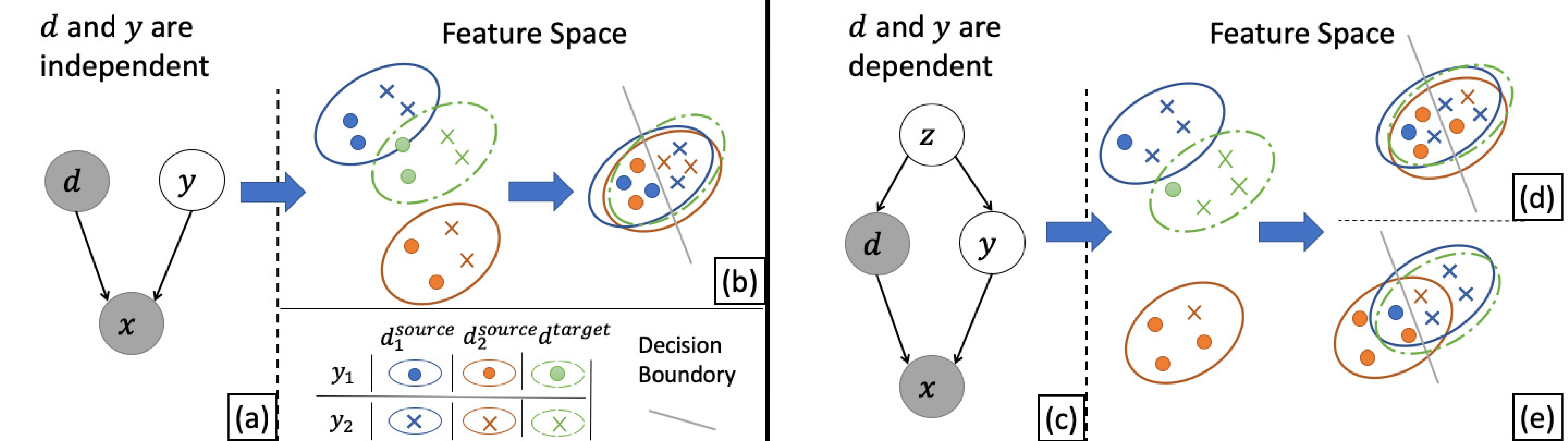}
  \caption{Explanation of domain-class dependency and the induced trade-off.
  (a) When the domain and the class are independent, (b) domain invariance and classification accuracy can be optimized at the same time, and the invariance prevents the classifier from overfitting to source domains.
  (c) When they are dependent, a trade-off exists between these two: (d) optimal classification accuracy cannot be achieved when perfect invariance is achieved, and (e) vice versa.
  We propose a method to lead explicitly to (e) rather than (d), because the primary purpose for domain generalization is classification, not domain-invariance itself.}
  \label{image:motivation}
  \vspace{-0.5cm}
\end{figure}

This paper considers DG under the situation where domain $d$ and class labels $y$ are statistically dependent owing to some common latent factor $z$ (Figure \ref{image:motivation}-(c)), which we referred to as {\it domain-class dependency}.
For example, the WISDM Activity Prediction dataset \cite{Kwapisz+2011WISDM}, where classes and domains correspond to activities and wearable device users, exhibits this dependency because of the (1) {\it data characteristics}: some activities (jogging and climbing stairs) are strenuous to the extent that some unathletic subjects avoided them, and (2) {\it data-collection errors}: other activities were added only after the study began and the initial subjects could not perform them.
Note that the dependency is common in real-world datasets and a similar setting has been investigated in DA studies \cite{Zhang+2013,Gong+2016}, but most prior DG studies overlooked the dependency; moreover, we need to follow a approach separate from DA because DG methods cannot require any access to target data, as we discuss further in Sec. \ref{related-work}.

% 代表的なドメイン汎化の手法として，特徴量を近づける系の手法があったことの指摘．
Most prior DG methods utilize invariant feature learning (IFL) \cite{Muandet+2013,Erfani+2016_sensor,Ghifary+2017_SCA,Xie+2017}, which can be negatively affected by the dependency.
IFL attempts to learn latent representation $h$ from input data $x$ which is invariant to domains $d$, or match multiple source domain distributions in feature space.
When source and target domains have some common structure (see, \cite{Muandet+2013}), matching multiple source domains leads to match source and target ones and thereby prevent the classifier from overfitting to source domains (Figure \ref{image:motivation}-(b)).
However, under the dependency, merely imposing the perfect domain invariance (which means $h$ and $d$ are independent) adversely affects the classification accuracy as pointed out by Xie et al. \cite{Xie+2017} and illustrated in Figure \ref{image:motivation}.
Intuitively speaking, since $y$ contains information about $d$ under the dependency, encoding information about $d$ into $h$ helps to predict $y$; however, IFL attempts to remove all domain information from $h$, which causes the trade-off.
Although that trade-off occurs in source domains (because we use only source data during optimization), it can also negatively affect the classification performance for target domains.
For example, if the target domain has characteristics similar (or same as an extreme case) to those of a certain source domain, giving priority to domain invariance obviously interferes with the DG performance (Figure \ref{image:motivation}-(d)).

% この論文の内容
In this paper, considering that prioritizing domain invariance under the trade-off can negatively affect the DG performance, we propose to maximize domain invariance within a range that does not interfere with the classification accuracy (Figure \ref{image:motivation}-(e)).
We first expand the analysis by \cite{Xie+2017} about domain adversarial nets (DAN), a well-used IFL method, and derive Theorem \ref{theo:trade-off-1} and \ref{theo:trade-off-2} which show the conditions under which domain invariance harms the classification accuracy.
In Theorem \ref{theo:desirable} we show that {\em accuracy-constrained domain invariance}, which we define as the maximum $H(d|h)$ ($H$ denotes entropy) value within a range that does not interfere with accuracy, equals $H(d|y)$.
In other words, when $H(d|h)=H(d|y)$, i.e., the learned representation $h$ contains as much domain information as the class labels, it does not affect the classification performance.
After deriving the theorems, we propose a novel method {\it adversarial feature learning with accuracy constraint (AFLAC)}, which leads to that invariance on adversarial training.
Empirical validations show that the performance of AFLAC is superior to that of baseline methods, supporting the importance of considering domain-class dependency and the efficacy of the proposed approach for overcoming the issue.

% この論文の新規性
The main contributions of this paper can be summarized as follows.
Firstly, we show that the implicit assumption of previous IFL methods, i.e., domain and class are statistically independent, is not valid in many real-world datasets, and it degrades the DG performance of them.
Secondly, we theoretically show to what extent latent representation can become invariant to domains without interfering with classification accuracy.
This is significant because the analysis guides the novel regularization approach that is suitable for our situation.
Finally, we propose a novel method which improves domain invariance while maintaining classification performance, and it enjoys higher accuracy than the IFL methods on both synthetic and three real-world datasets.

\section{Preliminary and Related Work}

\subsection{Problem Statement of Domain Generalization}

Denote ${\cal X}, {\cal Y}$, and ${\cal D}$ as the input feature, class label, and domain spaces, respectively.
With random variables $x \in {\cal X}, \; y \in {\cal Y}$, $d \in {\cal D}$，we can define the probability distribution for each domain as $p(x, y|d)$.
For simplicity this paper assumes that $y$ and $d$ are discrete variables.
In domain generalization, we are given a training dataset consisting of $\{ x_i^s, y_i^s \}_{i=1}^{n^s} $ for all $s \in \{1, 2, ..., m\}$, where each $\{ x_i^s, y_i^s \}_{i=1}^{n^s}$ is drawn from the source domain $p(x, y|d=s)$.
Using the training dataset, we train a classifier $g: {\cal X} \to {\cal Y}$, and use the classifier to predict labels of samples drawn from unknown target domain $p(x, y|d=t)$.

\subsection{Related Work}\label{related-work}

% 特徴量を近づける系の中でも，E2Eの手法の紹介．
DG has been attracting considerable attention in recent years \cite{Muandet+2013,Shankar+2018}.
\cite{Li+ICCV2017} showed that non-end-to-end DG methods such as DICA \cite{Muandet+2013} and MTAE \cite{Ghifary+2015_ICCV} do not tend to outperform vanilla CNN, thus end-to-end methods are desirable.
End-to-end methods based on domain invariant representation can be divided into two categories: adversarial-learning-based methods such as DAN \cite{Ganin+2016,Xie+2017} and pre-defined-metric-based methods \cite{Ghifary+2017_SCA,Li+2018_CVPR}.

In particular, our analysis and proposed method are based on DAN, which measures the invariance by using a domain classifier (also known as a discriminator) parameterized by deep neural networks and imposes regularization by deceiving it.
Although DAN was originally invented for DA, \cite{Xie+2017} demonstrated its efficacy in DG.
In addition, they intuitively explained the trade-off between classification accuracy and domain invariance, but did not suggest any solution to the problem except for carefully tuning a weighting parameter.
AFLAC also relates to domain confusion loss \cite{Tzeng+2015_ddg} in that their encoders attempted to minimize Kullback-Leibler divergence (KLD) between the output distribution of the discriminators and some domain distribution ($p(d|y)$ in AFLAC and uniform distribution in \cite{Tzeng+2015_ddg}), rather than to deceive the discriminator as DAN.
% Other researchers \cite{Louppe+2017_pivot} provided an analysis similar to that of \cite{Xie+2017}.
% However, their analysis differed in that they focused on the relation between the nuisance variables (domains) and output distribution of the domain classifier.

% 既存のドメイン汎化のSemantic Alignmentの説明と関連性を説明
Several studies that address DG without utilizing IFL have been conducted.
For example, CCSA \cite{motiian+2017CCSA}, CIDG \cite{Li+2018(CIDG)}, and CIDDG \cite{Li_2018_ECCV} proposed to make use of semantic alignment, which attempts to make latent representation given class label ($p(h|y)$) identical within source domains.
This approach was originally proposed by \cite{Gong+2016} in the DA context, but its efficacy to overcome the trade-off problem is not obvious.
Also, CIDDG, which is the only adversarial-learning-based semantic alignment method so far, needs the same number of domain classification networks as domains whereas ours needs only one.
\cite{Zhao+2017} also proposed a variant of adversarial-learning-based IFL method similar to ours, i.e., their method is also intended to maximize domain-invariance without affecting classification performance.
Although their method needs to estimate true data distribution $p(y|x)$ with DNN, ours only needs to estimate $p(d|y)$, which is easily conducted when $y$ and $d$ are discrete random variable.
CrossGrad \cite{Shankar+2018}, which is one of the recent state-of-the-art DG methods, utilizes data augmentation with adversarial examples.
However, because the method relies on the assumption that $y$ and $d$ are independent, it might not be directly applicable to our setting.
MLDG \cite{Li+2018_MLDG}, MetaReg \cite{Balaji+2018_metareg}, and Feature-Critic \cite{Li+2019_hdg}, other state-of-the-art methods, are inspired by meta-learning.
These methods make no assumption about the relation between $y$ and $d$; hence, they could be combined with our proposed method in principle.

As with our paper, \cite{Li+2018(CIDG),Li_2018_ECCV} also pointed out the importance of considering the types of distributional shifts that occur, and they address the shift of $p(y|x)$ across domains caused by the causal structure $y \to x$.
However, the causal structure does not cause the trade-off problem as long as $y$ and $d$ are independent (Figure \ref{image:motivation}-(a, b)), thus it is essential to consider and address domain-class dependency problem.
They also proposed to correct the domain-class dependency with the class prior-normalized weight, which enforces the prior probability for each class to be the same across domains.
Its motivation is different from ours in that it is intended to avoid overfitting whereas we address the trade-off problem.
% ; moreover, a reweighting approach such as this can remove domain information that is useful in the classification task.

In DA, \cite{Zhang+2013,Gong+2016} address the situation where $p(y)$ changes across the source and target domains by correcting the change of $p(y)$ using unlabeled target domain data, which is often accomplished at the cost of classification accuracy for the source domain.
However, this approach is not applicable (or necessary) to DG because we are agnostic on target domains and cannot run such adaptation step in DG.
Instead, this paper is concerned with the change of $p(y)$ within source domain and proposes to maximize the classification accuracy for source domains while improving the domain invariance.

It is worth mentioning that IFL has been used for many other context other than DG, e.g., DA \cite{Tzeng+2014,Ganin+2016}, domain transfer \cite{Lample_2017_fader,Chou+2018VC}, and fairness-aware classification \cite{Zemel+2013_fair,Louizos+2016_VFAE,Madras+2018_Fair}.
However, adjusting it to each specific task is likely to improve performance.
For example, in the fairness-aware classification task \cite{Madras+2018_Fair} proposed to optimize the fairness criterion directly instead of applying invariance to sensitive variables.
By analogy, we adapted IFL for DG so as to address the domain-class dependency problem.

\section{Our approach}

\subsection{Domain Adversarial Networks} \label{ss:DAN}

% [Domain Adversarial Netsの紹介]
In this section, we provide a brief overview of DAN \cite{Ganin+2016}, on which our analysis and proposed method are based.
DAN trains a domain discriminator that attempts to predict domains from latent representation encoded by an encoder, while simultaneously training the encoder to remove domain information by deceiving the discriminator.

Formally, we denote $f_E(x), q_M(y|h),$ and $q_D(d|h)$ ($E, M$, and $D$ are their parameters) as the deterministic encoder, probabilistic model of the label classifier, and that of domain discriminator, respectively.
Then, the objective function of DAN is described as follows:
\begin{align}
    \min_{E,M} \max_{D} J(E, M, D) =& \mathbb{E}_{p(x, d, y)} [ - \gamma L_d + L_y ], \label{eq:DANN} \\
                              where \;\; L_d \coloneqq -& \log{q_D (d|h=f_E(x))}, \;\; L_y \coloneqq - \log{q_M(y|h=f_E(x))}. \nonumber
\end{align}
Here, the second term in Eq. \ref{eq:DANN} simply maximizes the log likelihood of $q_M$ and $f_E$ as well as in standard classification problems.
On the other hand, the first term corresponds to a minimax game between the encoder and discriminator, where the discriminator $q_D(d|h)$ tries to predict $d$ from $h$ and the encoder $f_E(x)$ tries to fool $q_D(d|h)$.

As \cite{Xie+2017} originally showed, the minimax game ensures that the learned representation has no or little domain information, i.e., the representation becomes domain-invariant.
This invariance ensures that the prediction from $h$ to $y$ is independent from $d$, and therefore hopefully facilitates the construction of a classifier capable of correctly handling samples drawn from unknown domains (Figure \ref{image:motivation}-(b)).
Below is a brief explanation.

Because $h$ is a deterministic mapping of $x$, the joint probability distribution $p(h, d, y)$ can be defined as follows:
\begin{align}
    p(h, d, y) =& \int_x p(x, d, h, y) dx = \int_x p(x, d, y) p(h|x) dx \nonumber \\
               =& \int_x p(x, d, y) \delta (f_E(x)=h)dx, \label{eq:tildep}
\end{align}
and in the rest of the paper, we denote $p(h, d, y)$ as $\tilde{p}_E (h, d, y)$ because it depends on the encoder's parameter $E$.
Using Eq. \ref{eq:tildep}, Eq. \ref{eq:DANN} can be replaced as follows:
\begin{align}
    \min_{E,M} & \max_{D} J(E, M, D) = \mathbb{E}_{\tilde{p}_E(h, d, y)} [ \gamma \log{q_D (d|h)} - \log{q_M(y|h)} ]. \label{eq:DANN2}
\end{align}
Assuming $E$ is fixed, the solutions $M^*$ and $D^*$ to Eq. \ref{eq:DANN2} satisfy $q_{M^*}(y|h)=\tilde{p}_E(y|h)$ and $\; q_{D^*}(d|h)=\tilde{p}_E(d|h)$.
Substituting $q_{M^*}$ and $q_{D^*}$ into Eq. \ref{eq:DANN2} enable us to obtain the following optimization problem depending only on $E$:
\begin{align}
    \min_{E} J(E) =& - \gamma H_{\tilde{p}_E}(d|h) + H_{\tilde{p}_E}(y|h). \label{eq:DANN_ent}
\end{align}
Solving Eq. \ref{eq:DANN_ent} allows us to obtain the solutions $M^*$, $D^*$, and $E^*$, which are in Nash equilibrium.
Here, $H_{\tilde{p}_E}(d|h)$ means conditional entropy with the joint probability distribution $\tilde{p}_E(d, h)$.
Thus, minimizing the second term in Eq. \ref{eq:DANN_ent} intuitively means learning (the mapping function $f_E$ to) the latent representation $h$ which contains as much information about $y$ as possible.
On the other hand, the first term can be regarded as a regularizer that attempts to learn $h$ that is invariant to $d$.

\subsection{Trade-off Caused by Domain-Class Dependency} \label{ss:DAN_trade_off_analysis}

Here we show that the performance of DAN is impeded by the existence of domain-class dependency.
Concretely, we show that the dependency causes the trade-off between classification accuracy and domain invariance: when $d$ and $y$ are statistically dependent, no values of $E$ would be able to optimize the first and second term in Eq. \ref{eq:DANN_ent} at the same time.
Note that the following analysis also suggests that most IFL methods are negatively influenced by the dependency.

To begin with, we consider only the first term in Eq. \ref{eq:DANN_ent} and address the optimization problem:
\begin{align}
    \min_{E} J_1(E) =& - \gamma H_{\tilde{p}_E}(d|h). \label{eq:DANN_ent_1}
\end{align}
Using the property of entropy, $H_{\tilde{p}_E}(d|h)$ is bounded:
\begin{align}
     H_{\tilde{p}_E}(d|h) \leq H(d). \label{eq:entropy_bound_1}
\end{align}
Thus, Eq. \ref{eq:DANN_ent_1} has the solution $E^{*}_1$ which satisfies the following condition:
\begin{align}
    H_{\tilde{p}_{E^{*}_1}} (d|h) = H(d). \label{complete_invariance}
\end{align}
Eq. \ref{complete_invariance} suggests that the regularizer in DAN is intended to remove all information about domains from latent representation $h$, thereby ensuring the independence of domains and latent representation.

Next, we consider only the second term in Eq. \ref{eq:DANN_ent}, thereby addressing the following optimization problem:
\begin{align}
    \min_{E} J_2(E) = H_{\tilde{p}_E}(y|h). \label{eq:DANN_ent_2}
\end{align}
Considering $h$ is the mapping of $x$, i.e., $h=f_E(x)$, the solution $E^{*}_2$ to Eq. \ref{eq:DANN_ent_2} satisfies the following equation:
\begin{align}
    H_{\tilde{p}_{E^{*}_2}}(y|h) = H(y|x).  \label{eq:entropy_bound_2}
\end{align}

Here we obtain $E^{*}_1$ and $E^{*}_2$, which can achieve perfect invariance and optimal classification accuracy, respectively.
Using them, we can obtain the following theorem, which shows the existence of the trade-off between invariance and accuracy: perfect invariance ($E^{*}_1$) and optimal classification accuracy ($E^{*}_2$) cannot be achieved at the same time.

\begin{theo} \label{theo:trade-off-1}
    When $H(y|x)=0$ ,i.e., there is no labeling error, and $H(d) > H(d|y)$, i.e., the domain and class are statistically dependent, $E^{*}_1 \neq E^{*}_2$ holds.
\end{theo}

\begin{proo} \label{proo:trade-off-1}
    Assume $E^{*}_1=E^{*}_2=E^{*}$.
    Using the properties of entropy, we can obtain the following:
    \begin{align}
        % H_{\tilde{p}_{E}}(d|h) - H(d|y) \leq H_{\tilde{p}_{E}}(y|h) \label{eq:entropy_bound_3}
        H_{\tilde{p}_{E}}(d|h) \leq H_{\tilde{p}_{E}}(d, y|h) = H_{\tilde{p}_{E}}(d |h, y) + H_{\tilde{p}_{E}}(y|h) \leq H_{\tilde{p}_{E}}(d |y) + H_{\tilde{p}_{E}}(y|h). \label{eq:entropy_bound_3}
    \end{align}
    Substituting $H_{\tilde{p}_{E^*}}(y|h)=H(y|x)$ and $H_{\tilde{p}_{E^*}}(d|h)=H(d)$ into Eq. \ref{eq:entropy_bound_3}, we can obtain the following condition:
    \begin{align}
        H(d) - H(d|y)  \leq  H(y|x). \label{eq:entropy_bound_4}
    \end{align}
    Because the domain and class are dependent on each other, the following condition holds:
    \begin{align}
        0 < H(d) - H(d|y)  \leq H(y|x), \label{eq:entropy_bound_5}
    \end{align}
    but Eq. \ref{eq:entropy_bound_5} contradicts with $H(y|x)=0$.
    Thus, $E^{*}_1 \neq E^{*}_2$.
\end{proo}

Theorem \ref{theo:trade-off-1} shows that the domain-class dependency causes the trade-off problem.
Although it assumes $H(y|x)=0$ for simplicity, we cannot know the true value of $H(y|x)$ and there are many cases in which little or no labeling errors occur and thus $H(y|x)$ is close to $0$.

In addition, we can omit the assumption and obtain a more general result:
\begin{theo} \label{theo:trade-off-2}
    When $I(d;y) \coloneqq H(d) - H(d|y) > H(y|x)$, $E^{*}_1 \neq E^{*}_2$ holds.
\end{theo}

\begin{proo}
    Similar to Proof \ref{proo:trade-off-1}, we assume that $E^{*}_1=E^{*}_2$ and thus Eq. \ref{eq:entropy_bound_4} is obtained.
    Obviously, Eq. \ref{eq:entropy_bound_4} does not hold when $H(d) - H(d|y) > H(y|x)$.
\end{proo}

Theorem \ref{theo:trade-off-2} shows that when the mutual information of the domain and class $I(d; y)$ is greater than the labeling error $H(y|x)$, the trade-off between invariance and accuracy occurs.
Then, although we cannot know the true value of $H(y|x)$, the performance of DAN and other IFL methods are likely to decrease when $I(d; y)$ has large value.

\subsection{Accuracy-Constrained Domain Invariance} \label{ss:desirable-invariance}

If we cannot avoid the trade-off, the next question is to decide how to accommodate it, i.e., to what extent the representation should become domain-invariant for DG tasks.
Here we provide the notion of accuracy-constrained domain invariance, which is the maximum domain invariance within a range that does not interfere with the classification accuracy.
The reason for the constraint is that the primary purpose of DG is the classification for unseen domains rather than the invariance itself, and the improvement of the invariance could detrimentally affect the performance.
For example, in WISDM, if we know the target activity was performed by a young rather than an old man, we might predict the activity to be jogging with a higher probability; thus, we would have to avoid removing such domain information that may be useful in the classification task.

\begin{theo} \label{theo:desirable}
    Define accuracy-constrained domain invariance as the maximum $H_{\tilde{p}_E}(d|h)$ value under the constraint that $H(y|x) = 0$, i.e., there is no labeling error, and classification accuracy is maximized, i.e., $H_{\tilde{p}_{E}}(y|h) = H(y|x)$.
    Then, accuracy-constrained domain invariance equals $H(d|y)$.
\end{theo}

\begin{proo} \label{proo:desirable}
    Using Eq. \ref{eq:entropy_bound_3} and $H_{\tilde{p}_{E}}(y|h) = H(y|x)$, the following inequation holds:
    \begin{align}
        H_{\tilde{p}_{E}}(d|h) \leq H(y|x) + H(d|y). \label{eq:entropy_bound_6}
    \end{align}
    Substituting $H(y|x) = 0$ into Eq. \ref{eq:entropy_bound_6}, the following inequation holds:
        \begin{align}
        H_{\tilde{p}_{E}}(d|h) \leq H(d|y). \label{eq:entropy_bound_7}
    \end{align}
    Thus, the maximum $H_{\tilde{p}_E}(d|h)$ value under the optimal classification accuracy constraint is $H(d|y)$.
\end{proo}
Note that we could improve the invariance more when $H(y|x)>0$ (that is obvious considering Eq. \ref{eq:entropy_bound_6}), but we cannot know the true value of $H(y|x)$ as we discussed in Sec. \ref{ss:DAN_trade_off_analysis}.
Thus, accuracy-constrained domain invariance can be viewed as the worst-case gurantee.

\subsection{Proposed Method} \label{ss:proposed}

\begin{figure}[t]
  \centering
  \includegraphics[width=7.8cm]{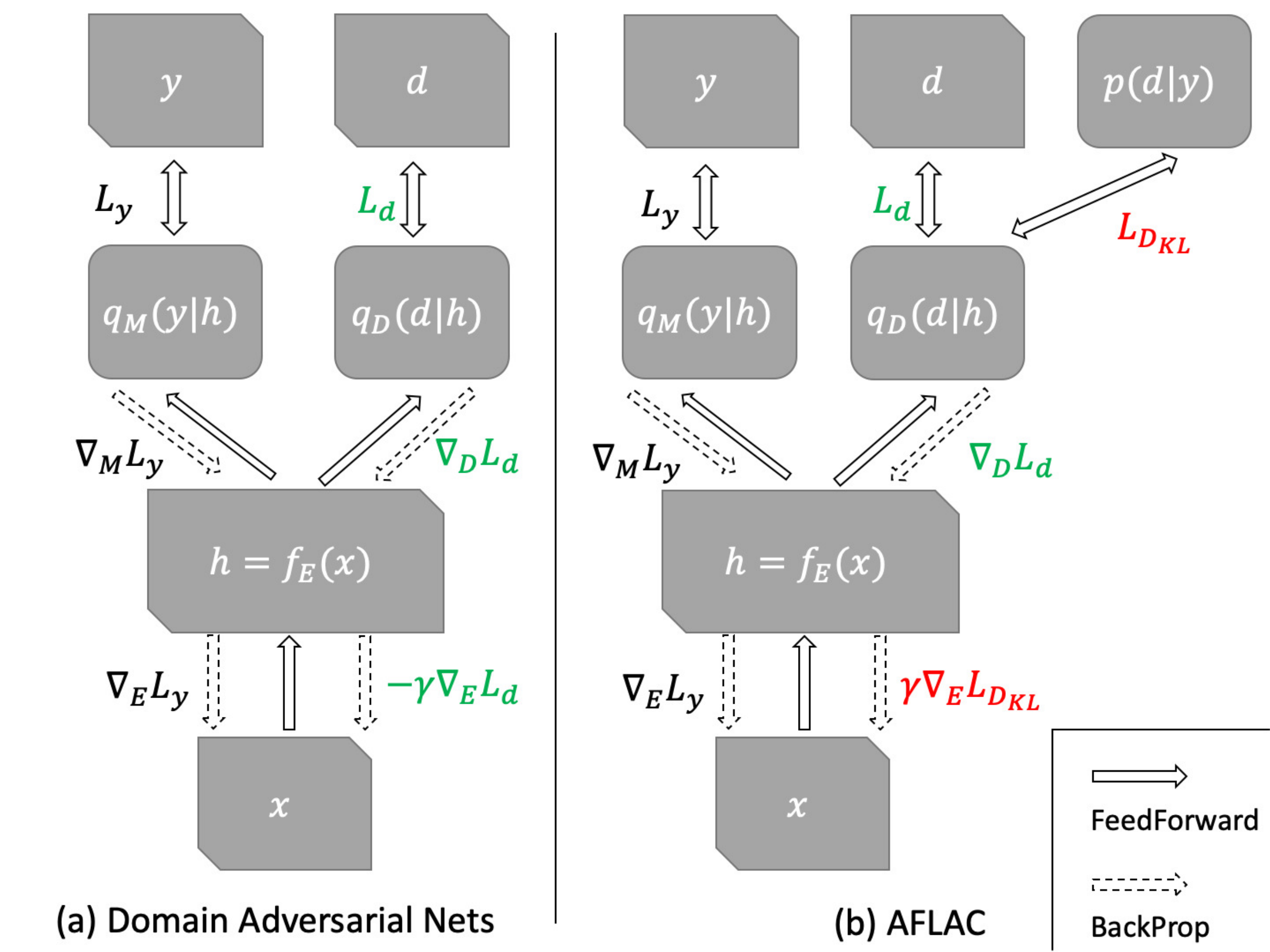}
  \vspace{-0.3cm} % 応急処置
  \caption{Comparative illustration of DAN and AFLAC.
  (a) The classifier and discriminator try to minimize $L_y$ and $L_d$, respectively. The encoder tries to minimize $L_y$ and maximize $L_d$ (fool the discriminator).
  (b) The discriminator tries to approximate true $\tilde{p}_{E}(d|h)$ by minimizing $L_d$. The encoder tries to minimize divergence between $\tilde{p}_{E}(d|h)$ and $p(d|y)$ by minimizing $L_{D_{KL}}$.
  }
  \label{image:aflac}
  \vspace{-0.5cm} % 応急処置
\end{figure}

Based on the above analysis, the remaining challenge is to determine how to achieve accuracy-constrained domain invariance, i.e., imposing regularization such that makes $H_{\tilde{p}_{E}}(d|h) = H(d|y)$ holds.
Although DAN might be able to achieve this condition by carefully tuning the strength of the regularizer ($\gamma$ in Eq. \ref{eq:DANN}), such tuning is time-consuming and impractical, as suggested by our experiments.
Alternatively, we propose a novel method named AFLAC by modifying the regularization term of DAN: whereas the encoder of DAN attempts to fool the discriminator, that of AFLAC attempts to directly minimize the KLD between $p(d|y)$ and $q_D(d|h)$.
Formally, AFLAC solves the following joint optimization problem by alternating gradient descent.
\begin{align}
    \min_{D}  W(E, D) = \mathbb{E}&_{p(x, d)} [ L_d ] \label{eq:IFLIC2} \\
    \min_{E,M} V(E, M) = \mathbb{E}&_{p(x, d, y)} [ \gamma L_{D_{KL}} + L_y ], \label{eq:IFLIC} \\
    where \;\;& L_{D_{KL}} \coloneqq D_{KL} [ p(d|y) | q_D(d|h = f_E(x))  ]. \nonumber
\end{align}
The minimization of $L_y$ and $L_d$, respectively, means maximization of the log-likelihood of $q_M$ and $q_D$ as well as in DAN.
However, the minimization of $L_{D_{KL}}$ differs from the regularizer of DAN in that it is intended to satisfy $q_D(d|h) = p(d|y)$.
And if $q_D(d|h)$ well approximates $\tilde{p}_{E}(d|h)$ by the minimization of $L_d$ in Eq. \ref{eq:IFLIC2}, the minimization of $L_{D_{KL}}$ leads to $\tilde{p}_{E}(d|h) = p(d|y)$.
Figure \ref{image:aflac}-(b) outlines the training of AFLAC.

Here we formally show that AFLAC is intended to achieve $H_{\tilde{p}_{E}}(d|h) = H(d|y)$ (accuracy-constrained domain invariance) by a Nash equilibrium analysis smilar to \cite{GANs,Xie+2017}.
As well as in Section \ref{ss:DAN}, $D^*$ and $M^*$, which are the solutions to Eqs. \ref{eq:IFLIC2}, \ref{eq:IFLIC} with fixed $E$, satisfy $q_D^* = \tilde{p}_{E}(d|h)$ and $q_M^* = \tilde{p}_{E}(y|h)$, respectively.
Thus, $V$ in Eq. \ref{eq:IFLIC} can be written as follows:
\begin{align}
    V(E) =  \mathbb{E} [ \gamma D_{KL} [ p(d|y) | \tilde{p}_E(d|h) ] ] + H_{\tilde{p}_{E}}(y|h). \label{eq:IFLIC_ent}
\end{align}
$E^*$, which is the solution to Eq. \ref{eq:IFLIC_ent} and in Nash equilibrium, satisfies not only $H_{\tilde{p}_{E^*}}(y|h)=H(y|x)$ (optimal classification accuracy) but also \\ $\mathbb{E}_{h, y \sim \tilde{p}_{E^*}(h, y)} [ D_{KL} [ p(d|y) | \tilde{p}_{E^*}(d|h) ] ] =0$, which is a sufficient condition for \\ $H_{\tilde{p}_{E^*}}(d|h) = H(d|y)$ by the definition of the conditional entropy.

In training, $p(x, d, y)$ in the objectives (Eqs. \ref{eq:IFLIC2}, \ref{eq:IFLIC}) is approximated by empirical distribution composed of the training data obtained from $m$ source domains, i.e., $\{ x_i^{(1)}, y_i^{(1)}, d=1 \}_{i=1}^{n^{(1)}},...,\{ x_i^{(m)}, y_i^{(m)}, d=m \}_{i=1}^{n^{(m)}}$.
Also, $p(d|y)$ used in Eq. \ref{eq:IFLIC} can be replaced by the maximum likelihood or maximum a posteriori estimator of it.
Note that, we could use some distances other than $D_{KL} [p(d|y) | q_D(d|h)]$ in Eq. \ref{eq:IFLIC}, e.g., $D_{KL} [q_D(d|h) | p(d|y)]$, but in doing so, we could not observe performance gain, hence we discontinued testing them.

\section{Experiments}

\subsection{Datasets}

Here we provide a brief overview of one synthetic and three real-world datasets (PACS, WISDM, IEMOCAP) used for the performance evaluation.
Although WISDM and IEMOCAP have not been widely used in DG studies, previous human activity recognition and speech emotion recognition studies (e.g., \cite{Ignatov+2017_ar,Etienne+2018_SER,Chen+2018_SER}) used them in the domain generalization setting (i.e., source and target domains are disjoint), so they can be regarded as the practical use case of domain generalization.
The concrete sample sizes for each $d$ and $y$, and the network architectures for each dataset are shown in supplementary.\footnote{Code and Supplementary are available at https://github.com/akuzeee/AFLAC}

{\bf BMNISTR} We created the Biased and Rotated MNIST dataset (BMNISTR) by modifying the sample size of the popular benchmark dataset MNISTR \cite{Ghifary+2015_ICCV}, such that the class distribution differed among the domains.
In MNISTR, each class is represented by 10 digits.
Each domain was created by rotating images by 15 degree increments: 0, 15, 30, 45, 60, and 75 (referred to as M0, ..., M75).
Each image was cropped to 16$\times$16 in accordance with \cite{Ghifary+2015_ICCV}.
We created three variants of MNISTR that have different types of domain-class dependency, referred to as BMNISTR-1 through BMNISTR-3.
As shown in Table \ref{table:mnistr}-left, BMNISTR-1, -2 have similar trends but different degrees of dependency, whereas BMNISTR-1 and BMNISTR-3 differ in terms of their trends.

{\bf PACS}
The PACS dataset \cite{Li+ICCV2017} contains 9991 images across 7 categories (dog, elephant, giraffe, guitar, house, horse, and person) and 4 domains comprising different stylistic depictions (Photo, Art painting, Cartoon, and Sketch).
It has domain-class dependency probably owing to the data characteristics.
For example, $p(y={\rm person}|d={\rm Phot})$ is much higher than $p(y={\rm person}|d={\rm Sketch})$, indicating that photos of a person are easier to obtain than those of animals, but sketches of persons are more difficult to obtain than those of animals in the wild.
For training, we used the ImageNet pre-trained AlexNet CNN \cite{Krizhevsky+2012_alex} as the base network, following previous studies \cite{Li+ICCV2017,Li+2018_MLDG}.
The two-FC-layer discriminator was connected to the last FC layer, following \cite{Ganin+2016}.

{\bf WISDM}
The WISDM Activity Prediction dataset contains sensor data of accelerometers of six human activities (walking, jogging, upstairs, downstairs, sitting, and standing) performed by 36 users (domains).
WISDM has the dependency for the reason noted in Sec. \ref{intro}.
In data preprocessing, we use the sliding-window procedure with 60 frames (=3 seconds) referring to \cite{Ignatov+2017_ar}, and the total number of samples was 18210.
We parameterized the encoder using three 1-D convolution layers followed by one FC layer and the classifier by logistic regression, following previous studies \cite{Yang+2015_HAR,Iwasawa2017_HAR}.

{\bf IEMOCAP}
The IEMOCAP dataset \cite{Busso+2008_iemocap} is the popular benchmark dataset for speech emotion recognition (SER), which aims at recognizing the correct emotional state of the speaker from speech signals.
It contains a total of 10039 utterances pronounced by ten actors (domains, referred to as Ses01F, Ses01M through Ses05F, Ses05M) with emotional categories, and we only consider the four emotional categories (happy, angry, sad, and neutral) referring to \cite{Chen+2018_SER,Etienne+2018_SER}.
Also, we refered to \cite{Chen+2018_SER} about data preprocessing: we split the speech signal into equal-length segments of 3s, and extracted 40-dimensional log Mel-spectrogram, its deltas, and delta-deltas.
We parameterized the encoder using three 2-D convolution layers followed by one FC layer and the classifier by logistic regression.

\subsection{Baselines}
To demonstrate the efficacy of the proposed method AFLAC, we compared it with vanilla CNN and adversarial-learning-based methods.
Specifically, {\bf (1) CNN} is a vanilla convolutional networks trained on the aggregation of data from all source domains. Although CNN has no special treatments for DG, \cite{Li+ICCV2017} reported that it outperforms many traditional DG methods.
{\bf (2) DAN } \cite{Xie+2017} is expected to generalize across domains utilizing domain-invariant representation, but it can be affected by the trade-off between domain invariance and accuracy as explained in Section \ref{ss:DAN_trade_off_analysis}.
{\bf (3) CIDDG} is our re-implementation of the method proposed in \cite{Li_2018_ECCV}, which is designed to achieve semantic alignment on adversarial training.
Additionally, we used {\bf (4) AFLAC-Abl}, which is a version of AFLAC modified for ablation studies.
AFLAC-Abl replaces $D_{KL} [ p(d|y) | q_D(d|h) ]$ in Eq. \ref{eq:IFLIC} of $D_{KL} [ p(d) |  q_D(d|h) ]$, thus it attempts to learn the representation that is perfectly invariant to domains or make $H(d|h)=H(d)$ hold as well as DAN.
Comparing AFLAC and AFLAC-Abl, we measured the genuine effect of taking domain-class dependency into account.
When training AFLAC and AFLAC-Abl, we cannot obtain true $p(d|y)$ and $p(d)$, hence we used their maximum likelihood estimators for calculating the KLD terms.

\subsection{Experimental Settings}

For all the datasets and methods, we used RMSprop for optimization.
Further, we set the learning rate, batch size, and the number of iterations as 5e-4, 128, and 10k for BMNISTR; 5e-5, 64, and 10k for PACS; 1e-4, 64, and 10k for IEMOCAP; 5e-4 (with exponential decay with decay step 18k and 24k, and decay rate 0.1), 128, and 30k for WISDM, respectively.
Also, we used the annealing of weighting parameter $\gamma$ proposed in \cite{Ganin+2016}, and unless otherwise mentioned chose $\gamma$ from $\{0.0001, 0.001, 0.01, 0.1, 1, 10\}$ for DAN, CIDDG, AFLAC-Abl, and AFLAC.
Specifically, on BMNISTR and PACS, we employed a leave-one-domain-out setting \cite{Ghifary+2015_ICCV}, i.e., we chose one domain as target and used the remaining domains as source data.
Then we split the source data into 80\% of training data and 20\% of validation data, assuming that target data are not absolutely available in the training phase.
On IEMOCAP, we chose the best $\gamma$ from \\ $\{0.0001, 0.001, 0.01, 0.1, 1, 10, 100, 1000\}$ using disjoint validation domain, referring to \cite{Etienne+2018_SER,Chen+2018_SER}.
On WISDM, we randomly selected  $<$20 / 16$>$ users as $<$source / target$>$ domains, and split the source data into training and validation data because one-domain-leave-out evaluation is computationally expensive.
Then, we conducted experiments multiple times with different random weight initialization; we trained the models on 10, 20, and 20 seeds in BMNISTR, WISDM, and IEMOCAP, chose the best hyperparameter that achieved the highest validation accuracies measured in each epoch, and reported the mean scores (accuracies and F-measures) for the hyperparameter.
On PACS, because it requires a long time to train on, we chose the best $\gamma$ from $\{0.0001, 0.001, 0.01, 0.1\}$ after three experiments, and reported the mean scores in experiments with 15 seeds.

\begin{table}[t]
  \caption{
  Left: Sample sizes for each domain-class pair in BMNISTR.
  Those for the classes 0$\sim$4 are variable across domains, whereas the classes 5$\sim$9 have identical sample sizes across domains.
  Right: Mean F-measures for the classes 0$\sim$4 and classes 5$\sim$9 with the target domain M0. RI denotes relative improvement of AFLAC to AFLAC-Abl}
  \vspace{-0.3cm} % 応急処置
  \label{table:mnistr}

  \begin{tabular}{cc}
  \begin{minipage}{0.5\hsize}
    \centering
    \scalebox{0.80}{
      \begin{tabular}{r|r|rrrrrr}
        \hline
        Dataset & Class & M0 & M15 & M30 & M45 & M60 & M75 \\
        \hline
        BMNISTR-1 & 0$\sim$4 & 100 & 85 & 70 & 55 & 40 & 25 \\
                  & 5$\sim$9 & 100 & 100 & 100 & 100 & 100 & 100 \\
        BMNISTR-2 & 0$\sim$4 & 100 & 90 & 80 & 70 & 60 & 50 \\
                  & 5$\sim$9 & 100 & 100 & 100 & 100 & 100 & 100 \\
        BMNISTR-3 & 0$\sim$4 & 100 & 25 & 100 & 25 & 100 & 25 \\
                  & 5$\sim$9 & 100 & 100 & 100 & 100 & 100 & 100 \\
        \hline
      \end{tabular}
    }
  \end{minipage}
    \begin{minipage}{0.5\hsize}
    \centering
    \scalebox{0.65}{
      \begin{tabular}{r|r|rrrrrc}
      \toprule
                &       &    CNN &    DAN &  CIDDG &  AFLAC &  AFLAC & RI \\
      Dataset   & Class &        &        &        &   -Abl &        &    \\
      \midrule
      BMNISTR-1 & 0$\sim$4 &  83.86 &  84.54 &  87.50 &      87.46 & {\bf 90.62 }&                 3.6\% \\
                & 5$\sim$9 &  83.90 &  85.24 &  87.46 &      86.46 & {\bf 88.10 }&                 1.9\% \\
      BMNISTR-2 & 0$\sim$4 &  82.54 &  85.30 &  87.64 &      88.60 & {\bf 89.64 }&                 1.2\% \\
                & 5$\sim$9 &  82.18 &  85.80 &  86.74 &      87.60 & {\bf 89.04 }&                 1.6\% \\
      BMNISTR-3 & 0$\sim$4 &  71.26 &  79.22 &  76.76 &      76.56 & {\bf 80.02 }&                 4.5\% \\
                & 5$\sim$9 &  78.62 & {\bf 83.14 }&  82.64 &      82.94 &  82.80 &                -0.2\% \\
      \bottomrule
      \end{tabular}
    }
  \end{minipage}
  \end{tabular}
  \vspace{-0.5cm}
\end{table}

\subsection{Results}

We first investigated the extent to which domain-class dependency affects the performance of the IFL methods.
In Table \ref{table:mnistr}-right, we compared the mean F-measures for the classes 0 through 4 and classes 5 through 9 in BMNISTR with the target domain M0.
Recall that the sample sizes for the classes 0$\sim$4 are variable across domains, whereas the classes 5$\sim$9 have identical sample sizes across domains (Table \ref{table:mnistr}-left).
The F-measures show that AFLAC outperformed baselines in most dataset-class pairs, which supports that domain-class dependency reduces the performance of domain-invariance-based methods and that AFLAC can mitigate the problem.
Further, the relative improvement of AFLAC to AFLAC-Abl is more significant for the classes 0$\sim$4 than for 5$\sim$9 in BMNISTR-1 and BMNISTR-3, suggesting that AFLAC tends to increase performance more significantly for classes in which the domain-class dependency occurs.
Moreover, the improvement is more significant in BMNISTR-1 than in BMNISTR-2, suggesting that the stronger the domain-class dependency is, the lower the performance of domain-invariance-based methods becomes.
This result is consistent with Theorem \ref{theo:trade-off-2}, which shows that the trade-off is likely to occur when $I(d; y)$ is large.
Finally, although the dependencies of BMNISTR-1 and BMNISTR-3 have different trends, AFLAC improved the F-measures in both datasets.

\begin{table}[t]
  \caption{Accuracies for each dataset and target domain.
   The $I(d; y)$ column is estimated from source domain data, which indicates the domain-class dependency.}
  \vspace{-0.3cm} % 応急処置
  \label{table:target-acc}
  \centering
  \scalebox{0.7}{
\begin{tabular}{llllllll}
\toprule
          &    & I(d; y) &       CNN &       DAN &     CIDDG & AFLAC-Abl &     AFLAC \\
Dataset & Target &         &           &           &           &           &           \\
\midrule
BMNISTR-1 & M0 &   0.026 &  83.9 $\pm$ 0.4 &  85.0 $\pm$ 0.4 &  87.4 $\pm$ 0.3 &  87.0 $\pm$ 0.4 & {\bf 89.3 $\pm$ 0.4} \\
          & M15 &   0.034 &  98.5 $\pm$ 0.2 &  98.5 $\pm$ 0.1 &  98.3 $\pm$ 0.2 &  98.3 $\pm$ 0.2 & {\bf 98.8 $\pm$ 0.1} \\
          & M30 &   0.037 &  97.5 $\pm$ 0.1 &  97.4 $\pm$ 0.1 &  97.4 $\pm$ 0.2 &  97.6 $\pm$ 0.1 & {\bf 98.3 $\pm$ 0.2} \\
          & M45 &   0.036 &  89.9 $\pm$ 0.9 &  90.2 $\pm$ 0.6 &  89.8 $\pm$ 0.5 &  92.8 $\pm$ 0.5 & {\bf 93.3 $\pm$ 0.6} \\
          & M60 &   0.030 &  96.7 $\pm$ 0.3 &  97.0 $\pm$ 0.2 &  97.2 $\pm$ 0.1 &  96.6 $\pm$ 0.2 & {\bf 97.4 $\pm$ 0.2} \\
          & M75 &   0.017 &  87.1 $\pm$ 0.5 &  87.3 $\pm$ 0.4 & {\bf 88.2 $\pm$ 0.3} &  87.7 $\pm$ 0.5 &  88.1 $\pm$ 0.4 \\
          \cline{2-8}
          & Avg &      &      92.3 &      92.6 &      93.1 &      93.3 &     {\bf 94.2} \\
\hline
BMNISTR-2 & Avg &      &      92.2 &      92.7 &      93.1 &      94.0 &     {\bf  94.5} \\
\hline
BMNISTR-3 & Avg &      &      90.6 &      91.7 &      91.4 &      91.6 &     {\bf  92.9} \\
\hline
PACS & photo &   0.102 &  82.2 $\pm$ 0.4 &  81.8 $\pm$ 0.4 &\;\;\;- &  82.5 $\pm$ 0.4 & {\bf 83.5 $\pm$ 0.3} \\
          & art\_painting &   0.117 &  61.0 $\pm$ 0.5 &  60.9 $\pm$ 0.5 &\;\;\;- &  62.6 $\pm$ 0.4 & {\bf 63.3 $\pm$ 0.3} \\
          & cartoon &   0.131 & {\bf 64.9 $\pm$ 0.5} & {\bf 64.9 $\pm$ 0.6} &\;\;\;- &  64.2 $\pm$ 0.3 & {\bf 64.9 $\pm$ 0.3} \\
          & sketch &   0.023 & {\bf 61.4 $\pm$ 0.5} & {\bf 61.4 $\pm$ 0.5} &\;\;\;- &  59.6 $\pm$ 0.7 &  60.1 $\pm$ 0.7 \\
          \cline{2-8}
          & Avg &         &      67.4 &      67.2 &  \;\;\;-  &      67.2 &     {\bf 68.0} \\
\hline
WISDM & 16 users &   0.181 &  84.0 $\pm$ 0.4 &  83.8 $\pm$ 0.3 & {\bf 84.4 $\pm$ 0.4} &  83.7 $\pm$ 0.3 & {\bf 84.4 $\pm$ 0.3} \\
\hline
IEMOCAP & Ses01F &   0.005 &  56.0 $\pm$ 0.7 &  60.1 $\pm$ 0.7 &\;\;\;- & {\bf 62.9 $\pm$ 0.5} &  60.4 $\pm$ 0.9 \\
          & Ses01M &         &  61.0 $\pm$ 0.3 &  63.5 $\pm$ 0.5 &\;\;\;- & {\bf 68.0 $\pm$ 0.5} &  66.1 $\pm$ 0.3 \\
          & Ses02F &   0.045 &  61.2 $\pm$ 0.5 &  60.4 $\pm$ 0.5 &\;\;\;- & {\bf 65.8 $\pm$ 0.5} &  64.2 $\pm$ 0.4 \\
          & Ses02M &         & {\bf 76.6 $\pm$ 0.4} &  47.2 $\pm$ 0.7 &\;\;\;- &  64.7 $\pm$ 1.7 &  74.3 $\pm$ 1.3 \\
          & Ses03F &   0.037 &  69.2 $\pm$ 0.9 & {\bf 71.9 $\pm$ 0.4} &\;\;\;- &  70.0 $\pm$ 0.6 &  70.1 $\pm$ 0.4 \\
          & Ses03M &         &  56.9 $\pm$ 0.4 & {\bf 57.3 $\pm$ 0.5} &\;\;\;- &  56.2 $\pm$ 0.4 &  56.8 $\pm$ 0.4 \\
          & Ses04F &   0.120 &  75.5 $\pm$ 0.5 &  75.5 $\pm$ 0.6 &\;\;\;- &  75.4 $\pm$ 0.6 & {\bf 75.7 $\pm$ 0.6} \\
          & Ses04M &         &  58.5 $\pm$ 0.5 &  57.4 $\pm$ 0.5 &\;\;\;- &  58.7 $\pm$ 0.5 & {\bf 59.2 $\pm$ 0.5} \\
          & Ses05F &   0.063 &  61.8 $\pm$ 0.4 &  62.4 $\pm$ 0.5 &\;\;\;- &  61.9 $\pm$ 0.3 & {\bf 63.4 $\pm$ 0.7} \\
          & Ses05M &         &  47.6 $\pm$ 0.3 &  46.9 $\pm$ 0.4 &\;\;\;- &  49.6 $\pm$ 0.4 & {\bf 49.9 $\pm$ 0.4} \\
          \cline{2-8}
          & Avg &         &      62.4 &      60.3 &   \;\;\;- &      63.3 &     {\bf 64.0} \\
\bottomrule
\end{tabular}
  }
  \vspace{-0.5cm}
\end{table}

Next we compared the mean accuracies (with standard errors) in both synthetic (BMNISTR) and real-world (PACS, WISDM, and IEMOCAP) datasets (Table \ref{table:target-acc}).
Note that the performance of our baseline CNN on PACS, WISDM, and IEMOCAP is similar but partly different from that reported in previous studies (\cite{Li_2018_ECCV}, \cite{Ignatov+2017_ar}, and \cite{Etienne+2018_SER}, respectively) probably because the DG performance strongly depends on validation methods and other implementation details as reported in many recent studies \cite{Ignatov+2017_ar,Etienne+2018_SER,Balaji+2018_metareg,Li+2019_hdg}.
Also, we trained CIDDG only on BMNISTR and WISDM due to computational resource constraint.
This table enables us to make the following observations.
{\bf(1)} Domain-class dependency in real-world datasets negatively affects the DG performance of IFL methods.
The results obtained on PACS (Avg) and WISDM showed that the vanilla CNN outperformed the IFL methods (DAN and AFLAC-Abl).
% ここをちゃんと説明するためには，1.不変性を上げすぎた方がいい場合もあること，2.その場合validationドメインに過学習する危険性が高まることを言わないと
Additionally, the results on IEMOCAP shows that AFLAC tended to outperform AFLAC-Abl when $I(d; y)$ had large values (in Ses04 and Ses05), which is again consistent with Theorem \ref{theo:trade-off-2}.  % これを書くならPACSについても議論しなきゃでめんどくさい
These results support the importance of considering domain-class dependency in real-world datasets.
{\bf(2)} AFLAC performed better than the baselines on all the datasets in average, except for CIDDG on WISDM.
Note that AFLAC is more parameter efficient than CIDDG as we noted in Sec. \ref{related-work}.
These results supports the efficacy of the proposed model to overcome the trade-off problem.

\begin{figure}[t]
  \begin{tabular}{cccc}
    \begin{minipage}[l]{0.25\hsize}
      \subfigure[BMNISTR-1, M0]{\includegraphics[width=3cm]{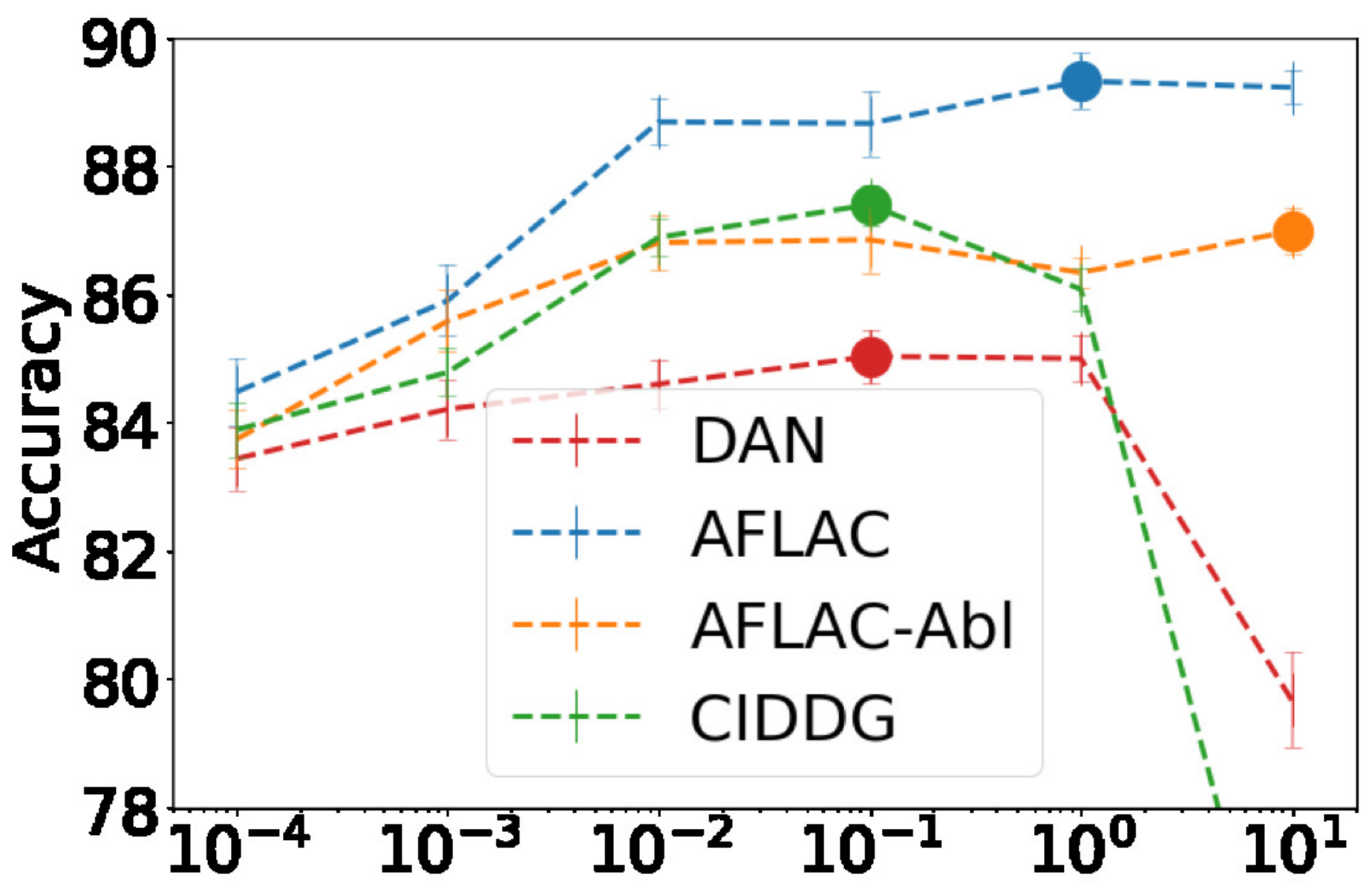}
      \label{image:bmnistr-0}}
    \end{minipage}
    \begin{minipage}[l]{0.25\hsize}
      \subfigure[WISDM]{\includegraphics[width=3cm]{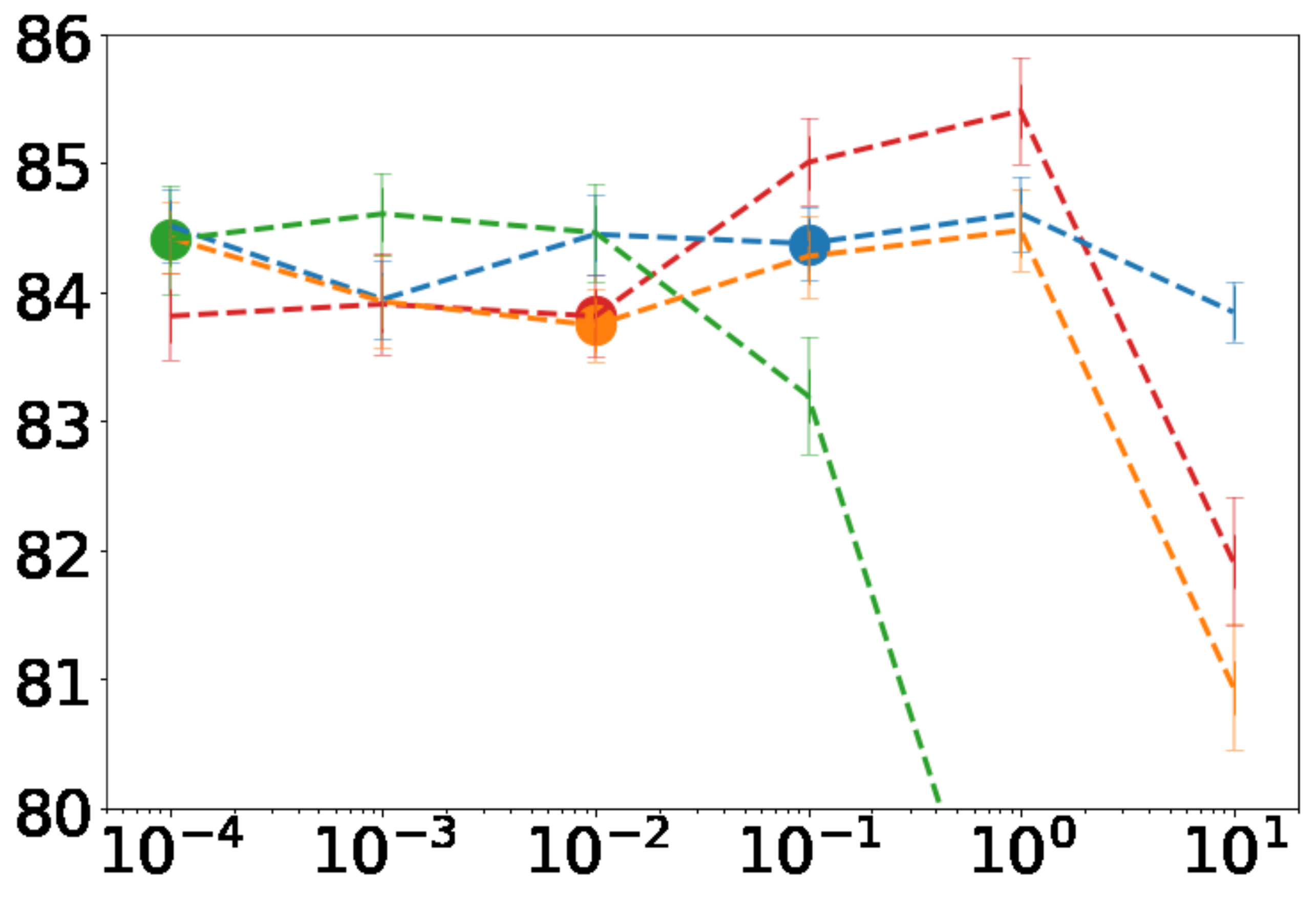}
      \label{image:bmnistr-75}}
    \end{minipage}
    \begin{minipage}[l]{0.25\hsize}
      \subfigure[IEMOCAP, 02M]{\includegraphics[width=3cm]{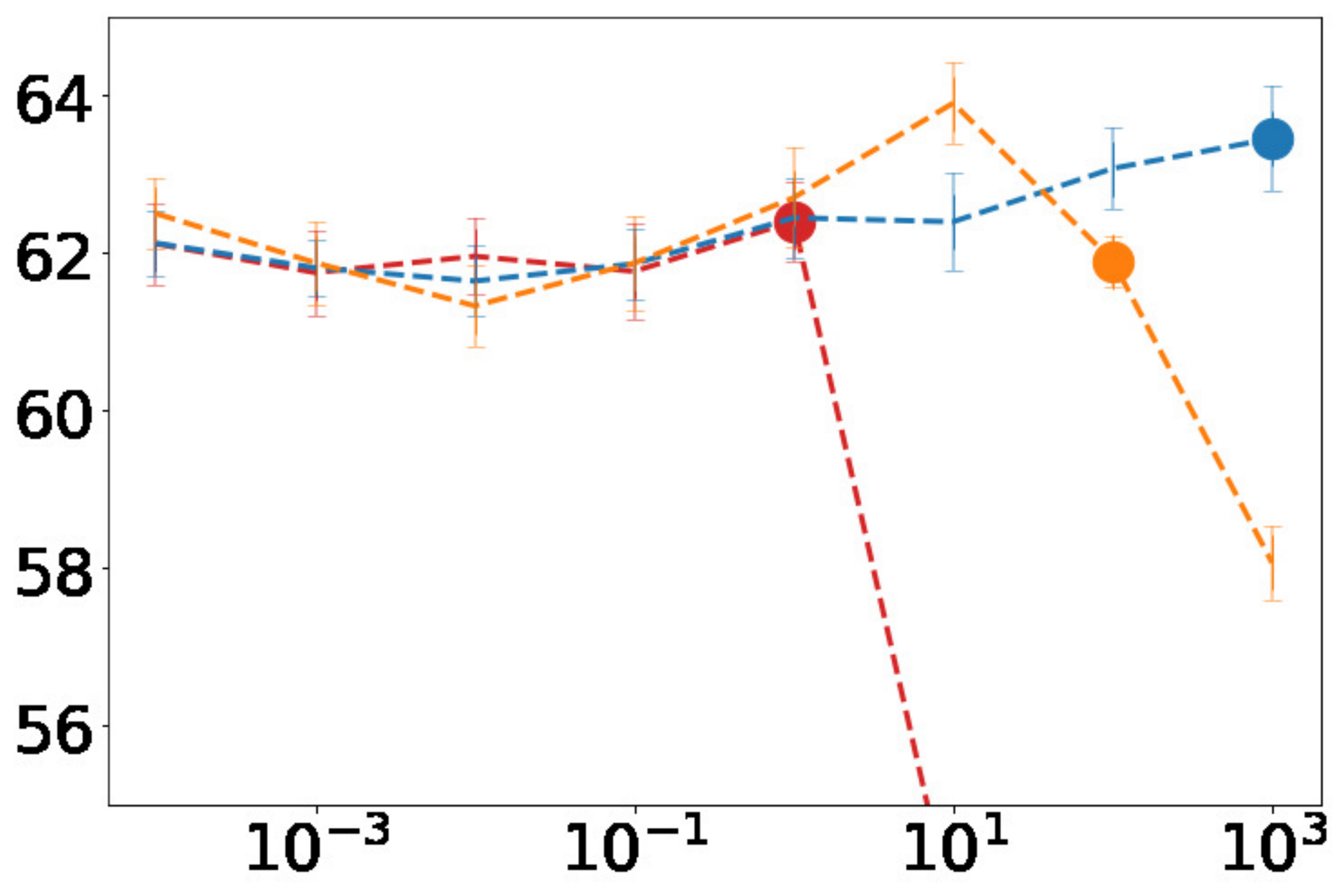}
      \label{image:wisdm-26}}
    \end{minipage}
    \begin{minipage}[l]{0.25\hsize}
      \subfigure[IEMOCAP, 05F]{\includegraphics[width=3cm]{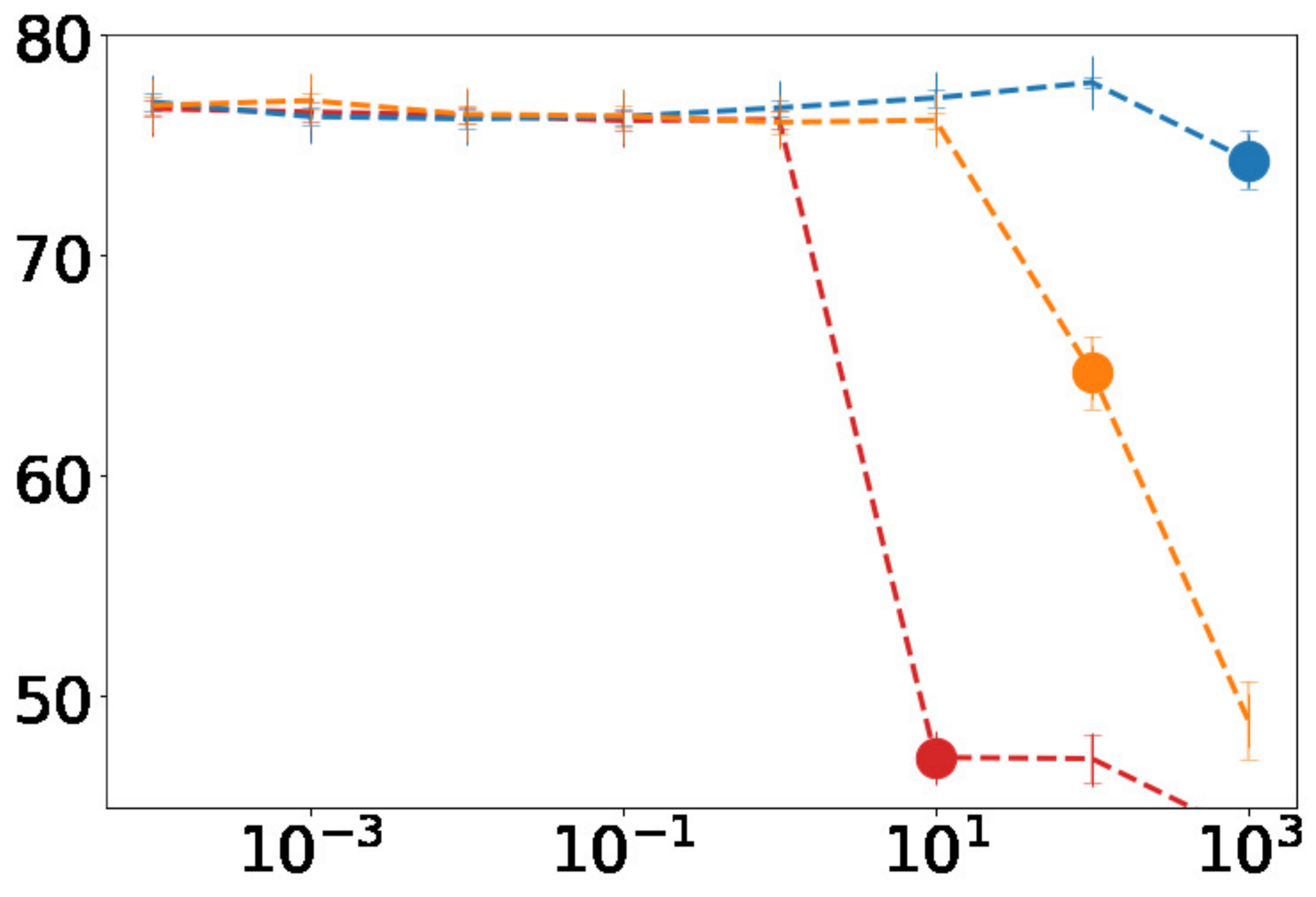}
      \label{image:wisdm-10}}
    \end{minipage}
  \end{tabular}
  \vspace{-0.4cm}
  \caption{Classification Accuracy with various $\gamma$. Each caption shows dataset name and target domain.
  The round markers correspond to $\gamma$ values chosen by validation.
  The error bars correspond to standard errors.}
  \label{image:hypara}
  \vspace{-0.4cm}
\end{figure}

Finally, we investigated the relationship between the strength of regularization and performance.
In DG, it is difficult to choose appropriate hyperparameters because we cannot use target domain data at valiadtion step (since they are not available during training); therefore, hyperparameter insensitivity is significant in DG.
Figure \ref{image:hypara} shows the hyperparameter sensitivity of the classification accuracies for DAN, CIDDG, AFLAC-Abl, and AFLAC.
These figures suggest that DAN and AFLAC-Abl sometimes outperformed AFLAC with appropriate $\gamma$ values, but there is no guarantee that such $\gamma$ values will be chosen by validation whereas AFLAC is robust toward hyperparameter choice.
Specifically, as shown in Figures \ref{image:hypara}-(b, d), DAN and AFLAC-Abl outperformed AFLAC with $\gamma=1$ and $10$, respectively.
One possible explanation of those results is that accuracy for target domain sometimes improves by giving priority to domain invariance at the cost of the accuracies for source domains, but AFLAC improves domain invariance only within a range that does not interfere with accuracy for source domains.
However, as shown in Figure \ref{image:hypara}, the performance of DAN and AFLAC-Abl are sensitive to hyperparameter choice.
For example, although they got high scores with $\gamma=1$ in Figure \ref{image:hypara}-(b), the scores dropped rappidly when $\gamma$ increases to $10$ or decreases to $0.01$.
Also, the scores of DAN and AFLAC-Abl in Figure \ref{image:hypara}-(c) dropped significantly with $\gamma > 10$, and such large $\gamma$ was indeed chosen by overfitting to validation domain.
On the other hand, Figures \ref{image:hypara}-(a, b, c, d) show that the accuracy gaps of AFLAC-Abl and AFLAC increase with strong regularization (such as when $\gamma=10$ or $100$).
These results suggest that AFLAC, as it was designed, does not tend to reduce the classification accuracy with strong regularizer, and such robustness of AFLAC might have yileded the best performance shown in Table \ref{table:target-acc}.
% {\bf(1)} The training of AFLAC and AFLAC-Abl tends to be more stable than that of DAN and CIDDG when the regularizer becomes strong.
% Figures \ref{image:hypara}-(a, b, c, d) show that AFLAC and AFLAC-Abl could achieve higher accuracies than DAN and CIDDG when $\gamma=1$ or $10$, i.e., the regularization is strong.
% This tendency might be because the regularizer of AFLAC is KLD and thus bounded by $0$, in contrast to that of DAN and CIDDG that can increase to infinity and destabilize the training.

\section{Conclusion}

In this paper, we addressed domain generalization under domain-class dependency, which was overlooked by most prior DG methods relying on IFL.
We theoretically showed the importance of considering the dependency and the way to overcome the problem by expanding the analysis of \cite{Xie+2017}.
We then proposed a novel method AFLAC, which maximizes domain invariance within a range that does not interfere with classification accuracy on adversarial training.
Empirical validations show the superior performance of AFLAC to the baseline methods, supporting the importance of the domain-class dependency in DG tasks and the efficacy of the proposed method to overcome the issue.
% Future work includes applying the regularization idea of $H(d|h)=H(d|y)$ to other methods, e.g., VAE\cite{Louizos+2016_VFAE} or CrossGrad because they have clear and tractable data generating process.


%
\end{document}